\documentclass[letterpaper, 10 pt, conference]{ieeeconf}  

\IEEEoverridecommandlockouts                              

\usepackage{graphics} 
\usepackage{epsfig} 
\usepackage{mathptmx} 
\usepackage{times} 
\usepackage{amsmath} 
\usepackage{amssymb}  
\usepackage{gensymb}
\usepackage[english]{babel}
\usepackage[utf8]{inputenc}

\providecommand{\etal}{{\em et~al.\ }}

\title{\LARGE \bf
Comparing View-Based and Map-Based Semantic Labelling \\ in Real-Time SLAM
}

\author{Zoe Landgraf$^{1}$, Fabian Falck$^{1}$, Michael Bloesch$^{1}$,  Stefan Leutenegger$^{2}$ and Andrew J. Davison$^{1}$
\thanks{Research presented in this paper has been supported by Dyson Technology Ltd.}
\thanks{$^{1}$Zoe Landgraf, Fabian Falck, Michael Bloesch and Andrew J. Davison are with the Dyson Robotics Laboratory, Department of Computing, Imperial College London, UK.
{\tt\small zoe.landgraf15@imperial.ac.uk}}%
\thanks{$^{2}$Stefan Leutenegger is with the Smart Robotics Lab, Department of Computing, Imperial College London, UK.}%
}

\begin{document}

\maketitle
\thispagestyle{empty}
\pagestyle{empty}

\begin{abstract}

Generally capable Spatial AI systems must build persistent scene representations where geometric models are combined with meaningful semantic labels. The many approaches to labelling scenes can be divided into two clear groups: \emph{view-based} which estimate labels from the input view-wise data and then incrementally fuse them into the scene model as it is built; and \emph{map-based} which label the generated scene model. However, there has so far been no attempt to quantitatively compare view-based and map-based labelling. Here, we present an experimental framework and comparison which uses real-time height map fusion as an accessible platform for a fair comparison, opening up the route to further systematic research in this area.


\end{abstract}

\section{INTRODUCTION}

As the cameras carried by robots or other smart devices move through scenes, their ability to operate and interact intelligently and persistently with their environments will depend on the quality of scene representation which they can build and maintain~\cite{Davison:ARXIV2018,Cadena:etal:TRO2016}. Convolutional Neural Networks (CNNs) have proven highly effective at semantic labelling and have been mainly used in two key approaches for scene labelling, each with its own set of advantages.

{\bf View-based labelling of raw input image data and incremental fusion of generated labels into the scene representation:} each camera frame is labelled and used together with the geometric correspondence available from a SLAM system to accumulate label estimates in the map. Up-to-date incremental map labels are thus available at frame rate with low latency. As each scene element is seen from multiple varying viewpoints, the independent image labellings contribute to robust fusion and the incremental correction of errors. The labelling CNN runs on ``raw'' data obtained directly from the camera, and can run at the sensor's natural resolution. Considering label correlations is computationally intractable and the individual pixel labels are thus generally assumed to be independent. Pixel-wise label fusion of every frame or keyframe during reconstruction offers a scalable solution to adding semantics to a variably sized scene, but can require unnecessary computation in areas which are easily segmented from one view. 

\begin{figure}[t]
\centering
\includegraphics[width=1.0\linewidth]{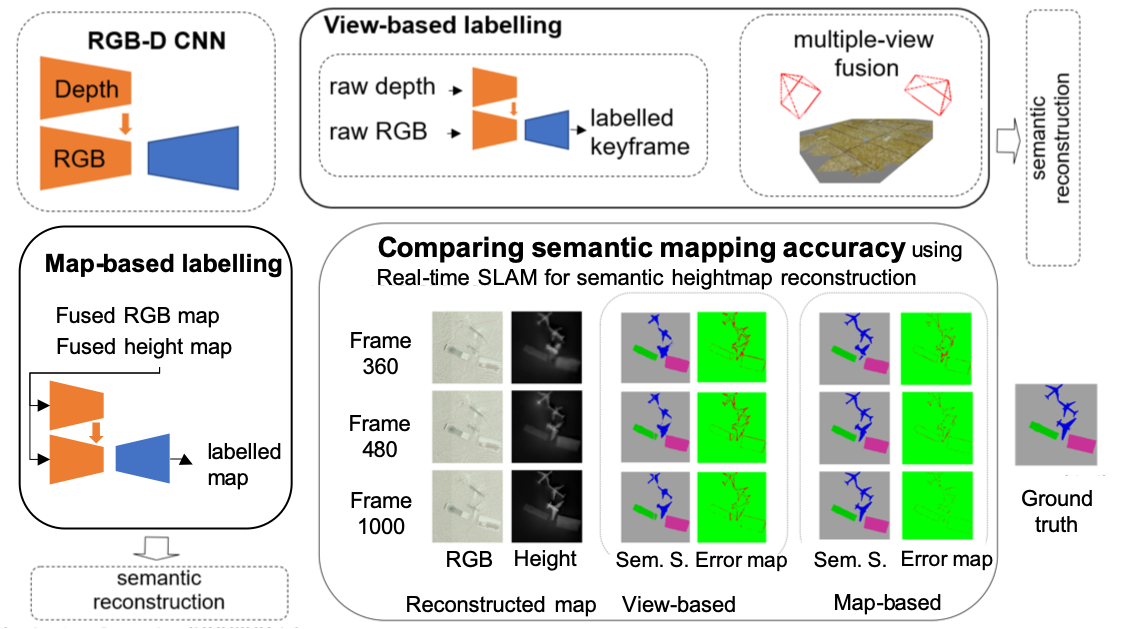}
\caption{Semantic labelling in SLAM is commonly done in one of two ways: incremental fusion of labels predicted for each view, or one-off labelling of the final reconstructed map. We compare these methods in controlled synthetic conditions in terms of accuracy, efficiency and robustness.}
\label{fig:teaser}
\vspace{2mm} \hrule
\end{figure}

{\bf One-off map-based labelling of the generated scene model:} a single labelling network such as a volumetric 3D CNN is applied to the whole reconstruction produced by a SLAM system, which contains both geometry and appearance information. This approach avoids the redundant work of labelling many overlapping input frames, and can be maximally efficient by operating on each scene element only once. Map-based methods can furthermore take advantage of global context over the whole scene when labelling each part and avoid the need for element-wise label fusion approaches which generally neglect label correlations. Finally, a CNN which learns and labels in canonical map space has much less scene variation to deal with due to rotation and scale changes. Its power will, however, be limited by the quality of the map reconstruction.

Semantic SLAM systems so far employ one or the other of the above approaches and sometimes attempt to combine both (see Section \ref{sec:related}). To our knowledge, the choice of approach is rarely based on quantitatively determined advantages for the particular use-case. We aim to establish a baseline for comparison of view-based and map-based approaches to open up systematic and quantitative research on their use within real-time SLAM. We have chosen table-top dense semantic object labelling as our experimental domain, using Height Map Fusion \cite{Zienkiewicz:etal:3DV2016} as our mapping approach. This deliberate choice of a system setup is more straightforward than general 3D geometry fusion, where there are many possible choices for 3D shape representation (e.g.\ point cloud, mesh, volumetric) and their corresponding network architectures for semantic labelling. Fused height maps can be labelled using a standard 2D CNN, allowing us to use networks with the same architecture for both view-based and map-based labelling, ensuring fairness in comparison with the aim that our results are transferable in their interpretation beyond the choices of a specific system. Our experiments are carried out on custom synthetic RGB-D data rendered with the SceneNet RGB-D methodology~\cite{McCormac:etal:ICCV2017}. We provide results which compare both labelling quality and computational load for view-based and map-based approaches and present the advantages of each, offering the potential for the design of optimal hybrid methods in the future.

\section{Related work}
\label{sec:related}
Preliminary work exists towards joint geometric and semantic SLAM (e.g.\ \cite{zhi2019scenecode}); yet, these systems are fairly limited in terms of accuracy and scaling.
Instead, the majority of state-of-the-art work relies on a sequential geometric reconstruction and frame-wise labelling, followed by semantic fusion. A few other approaches infer labels directly from the geometric 3D reconstruction and some combine both frame-wise and map-based methods. An overview of important work in each area is given below.

\subsection{Incremental View-Based Labelling and Fusion}
View-based approaches are generally based on well known image segmentation methods such as random forest classifiers \cite{Hermans:etal:ICRA2014, Stuckler:Behnke:JVCIR2014}, or depth segmentation based on normal angles \cite{tateno2015real} to generate semantic maps. McCormac \etal \cite{McCormac:etal:ICRA2017} introduced CNNs to incremental semantic fusion, obtaining semantic labels from RGB frames with a 2D CNN. Taking a slightly different direction, Ma \etal \cite{ma2017multi} propose a self-supervised multi-view prediction method and Xiang \etal \cite{Xiang:Fox:RSS2017} use a Recurrent Neural Network (RNN) to perform frame-wise segmentation of a sequence of RGB-D frames obtained from KinectFusion \cite{Newcombe:etal:ISMAR2011}. Recently, approaches such as PanopticFusion \cite{narita2019panopticfusion} and \cite{grinvald2019volumetric} extend incremental semantic fusion to object discovery and instance-aware maps.
Our view-based method leverages RGB-D images to segment 2D frames that are obtained in real-time from our SLAM system with a CNN. Similarly to \cite{McCormac:etal:ICRA2017}, we implement incremental fusion as a Bayesian update scheme, but tailor it to our reconstruction as described in Section \ref{sec:incremental_label_fusion}.

\subsection{Map based Labelling}
Labelling the 3D representation of a scene generally involves label inference from dense 3D representations such as TSDFs or voxel grids which are known to be expensive to process. Hence, approaches for labelling 3D representations have mainly been put forward for single objects \cite{Maturana:Schererl:IROS2015,Wu:etal:CVPR2015}. Recently, several have focused on representing and processing 3D data more efficiently \cite{WurmH:etal:IROS2011, riegler2017octnetfusion, wang2017cnn, yu2015multi}. Although most approaches focus on voxel grids, some approaches for classification of point clouds have been explored \cite{Qi:etal:CVPR2017}. Efficient 3D labelling at large scale, however, remains unsolved and only few have ventured into labelling a variably-sized reconstructed 3D scene. Landrieu \etal extended the idea of superpixels to 3D point clouds and proposed a superpoint method to label large scale LIDAR scans \cite{landrieu2018large}. Dai \etal \cite{Dai:etal:CVPR2018} proposed a fully convolutional, autoregressive, hierarchical coarse-to-fine 3D network to produce semantic labels together with geometry completion for a large 3D voxel grid scene. However, due to the expensive 3D nature of their input, the different levels of hierarchy in their network have to be trained separately. Roddick \etal \cite{roddick2018orthographic} project image features into an orthographic 3D space using a learned transformation, which removes the scale inconsistency, and creates a feature map with meaningful distances and without projective distortions of object appearance. They improve the efficiency of object detection from the orthographic map by collapsing voxel features along the vertical axis and then process the entire map of $80m \times 80m$ at a grid resolution of $0.5m$ at once. However, they do not address scalability in their method. In the map-based labelling approach applied in this paper, as in the methods above, we directly label the reconstruction. We employ a sliding window method dependent on our network's receptive field (see section \ref{sec:Map-based Labelling}) that allows scaling to arbitrarily large maps by avoiding potential GPU memory limitations when processing the entire map in a single forward pass. 

\subsection{Hybrid Methods}
Several methods employ a combination between map-based and view-based segmentation. Finman \etal \cite{finman2014efficient} incrementally segment a 3D point cloud using graph-based segmentation. They create small sub-parts of the map by joining the segmented parts with the border points of the existing map which they segment and fuse into the rest of the semantic map. Methods \cite{McCormac:etal:ICRA2017} and \cite{narita2019panopticfusion} combine view-based incremental labelling with a map-based smoothing step using a 3D Continuous Random Field. Vineet \etal \cite{vineet2015incremental} model their voxel grid with a volumetric, densely connected, pairwise Conditional Random Field and obtain semantic labels by first evaluating unary potentials in the image domain and then projecting them back onto the voxel grid. Dai \etal \cite{dai20183dmv} project 2D features extracted from multiple views onto a 3D voxel map, and use both geometry and projected features to predict per-voxel labels. 



\section{Method}
\label{sec:method}
The scenario for our experimental comparison is table-top reconstruction and semantic labelling of a scene containing scattered objects, selected from a number of \textit{ShapeNet} categories \cite{Shapenet:ARXIV2015}, as a depth camera browses the scene in an ad-hoc way. Since our focus is on a fundamental comparison of view-based and map-based labelling, we choose a height map representation for our scenes whose $2.5D$ nature allows us to use the same CNN network architecture designed for RGB-D input for both labelling methods. We use Height Map Fusion \cite{Zienkiewicz:etal:3DV2016} as our scene reconstruction backend. For our experiments, we opt for a synthetic environment based on rendered RGB-D data using the methodology from SceneNet RGB-D~\cite{McCormac:etal:ICCV2017}. A key reason for this decision is the need for a wide variety of scene configurations with semantic label ground truth in order to train high-performing view-based and map-based semantic segmentation networks. Furthermore, synthetic data gives us a high level of control over multiple experimental factors, such as the variety of viewpoints, noise, and ground truth for RGB, depth and camera poses. 


\subsection{Incremental Label Fusion}
\label{sec:incremental_label_fusion}
In the incremental fusion part of our study, we build on the real-time, multi-scale height map fusion system of Zienkiewicz~\etal~\cite{Zienkiewicz:etal:3DV2016}, by augmenting it with a semantic fusion capability. In~\cite{Zienkiewicz:etal:3DV2016}, the height map of a scene is modelled using a triangular mesh whose vertices have horizontal coordinates on a regular grid and associated variable heights which are estimated from incremental fusion. This system uses ORB-SLAM~\cite{Mur-Artal:etal:TRO2015} as a camera tracker, and geometry measurements can come from either a depth camera or incremental motion stereo in the pure monocular case. In our setup, which is based on synthetic data, we experiment with both ground truth as well as noisy camera poses and depth maps.

To add a semantic label fusion capability to the system, we associate a discrete distribution of semantic classes with every vertex of the mesh, and refine this distribution iteratively by projecting view-based semantic predictions onto the mesh in a per-surface-element-independent manner as in~\cite{McCormac:etal:ICRA2017}. For every vertex, only the pixels projected onto the adjacent faces contribute to the Bayesian update. 
We seek to compute the posterior distribution $P(v | M^{t})$ over semantic classes $v$ for a certain vertex, given projected measurements on adjacent faces $M^{t} = \{m^{1}, m^{2}, \dots, m^{t}\}$ for all timesteps $1, \ldots, t$. We define the measurement $m_{\boldsymbol{u}}^{t}$ as the network's prediction for a single pixel $\boldsymbol{u}$ given the image. We apply Bayes Rule to $P(v|M^{t})$ as follows:
\begin{equation}
    P(v|M^{t}) \propto P(m^{t}| M^{t-1}, v) P(v | M^{t-1})~. 
\end{equation}
We assume \textit{conditional independence} of the measurements given the vertex class, i.e. $P(m^{t}| M^{t-1}, v) = P(m^{t}|v)$, and can thus rewrite the above as:
\begin{equation}
{P(v| M^{t})} \propto P(m^{t}|v) {P(v| M^{t-1} )}
~,
    \label{eq:posteriorSemanticFusion}
\end{equation}
describing the relation between posterior $P(v| M^{t})$, measurement likelihood $P(m^{t}|v)$, and a-priori distribution $P(v| M^{t-1})$.
Note that we dropped the normalisation constant during the derivation and thus must normalise the posterior after evaluation.

For computational reasons, we also assume \textit{spatial independence} of the measurements and thus factorise the measurement likelihood into:
\begin{equation}
 P(m^{t} | v)  =\prod_{\mathbf{u} \in \mathcal{U}}{P(m_{\boldsymbol{u}}^{t}|v)}
    \label{eq:joint Likelihood}
~,
\end{equation}
where $\mathcal{U}$ denotes the set of pixels whose rays intersect with the surfaces adjacent to the given vertex and $P{(m_{\boldsymbol{u}}^{t}|v)}$ is the measurement likelihood at pixel $\boldsymbol{u}$ and time $t$ given the vertex class $v$. Since the projected measurement locations do not coincide with the location of the vertex, we model the measurement likelihood using a distance based decay $g(\cdot)$: 
\begin{eqnarray}
P{(m_{\boldsymbol{u}}^{t}|v)} &=& g(m_{\boldsymbol{u}}^{t}, v, d) \\
&=&  {\begin{cases}
       {\exp{(-\alpha d)}} a + b &\quad\text{if } m_{\boldsymbol{u}}^{t} = v \\
       \frac{1 - \exp{(-\alpha d)} a - b}{C-1} &\quad\text{if } m_{\boldsymbol{u}}^{t} \neq v
       
     \end{cases}
     }
     \label{eq:distance_based_decay}
~,
\end{eqnarray}
where $d$ is the Euclidean distance between the vertex and the projected pixel, $\alpha$ is a tuning parameter defining the decay rate and $a$ and $b$ are scaling factors based on the total number of semantic classes $C$ which ensure that $P{(m_{\boldsymbol{u}}^{t}|v)}$ models a uniform distribution as $d\to\infty$:
\begin{equation}
a = \frac{C-1}{C}, \ \ b = \frac{1}{C}
~.
\end{equation}
Intuitively, the closer the projected pixel is to the vertex, the more likely the pixel class is to coincide with the vertex class. The likelihood of a measurement being of any class other than the observed one is distributed uniformly.

Finally, the output of our network is not directly a measured class, but rather a distribution $m_{\boldsymbol{u}}^t(c)$ over possible classes $c$. This can be dealt with according to Bayes by evaluating a weighted average over classes:
\begin{equation}
\bar{g}(m_{\boldsymbol{u}}^{t}, v, d) = \sum_{c}{g(c, v, d) m_{\boldsymbol{u}}^t(c)}
~.
\end{equation}
where $\bar{g}(\cdot)$ replaces the measurement function for evaluating $P{(m_{\boldsymbol{u}}^{t}|v)}$ in Equation (\ref{eq:joint Likelihood}).

\subsection{Map-based Labelling}
\label{sec:Map-based Labelling}

\begin{figure}[t]
\begin{center}
   \includegraphics[width=1.0\linewidth]{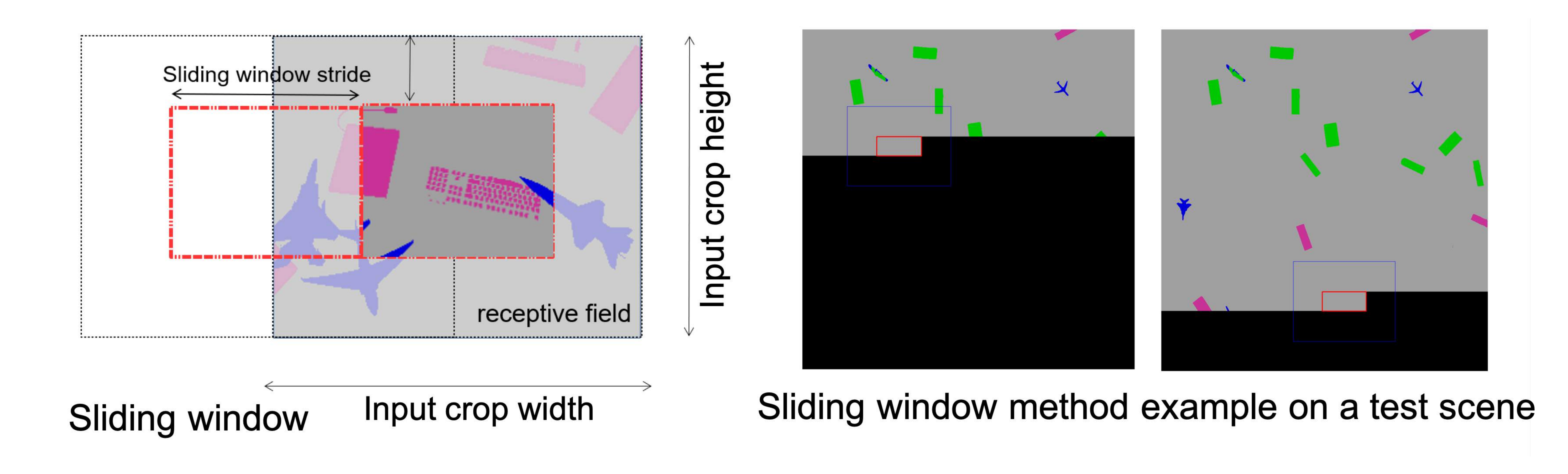}
\end{center}
   \caption{\textbf{Left:} Our map-based labelling uses a sliding window approach, with the offset determined according to the receptive field of the network. \textbf{Right:} Example of the sliding window method on one of our scenes.}
\label{fig:receptive_field_sliding_window_dummpy_graph}
\vspace{2mm} \hrule
\end{figure}

We model map-based labelling as a one-off segmentation of the entire reconstruction. In principle, labelling the scene directly could be implemented via a single pass through a very large CNN, but here, we propose to sequentially crop and segment parts of the map using a sliding window approach. Our choice makes our method scalable, as the sliding window can be applied to an arbitrarily sized height map without memory restrictions. While a naive sliding window approach of tiling the map and processing each sub-height-map would result in a loss of context in the border regions of each crop, we ensure correct segmentation by choosing the sliding window offset $o$ based on the theoretical receptive field $r$ of the network:
\begin{equation}
    o = d - 2r
    ~,
\end{equation}
where $d$ is the dimension of the network input in the sliding direction (width or height). Note that the sliding window offset is conservatively chosen and could practically be increased by considering the network's effective receptive field \cite{luo2016understanding}. With this method (see Figure \ref{fig:receptive_field_sliding_window_dummpy_graph}), we ensure the same context for every pixel as would be obtained during a single forward pass of the entire map through our network, while avoiding GPU memory limitations. 



\subsection{Scene Dataset Generation}

\begin{figure}[h]
\begin{center}
  \includegraphics[width=1.0\linewidth]{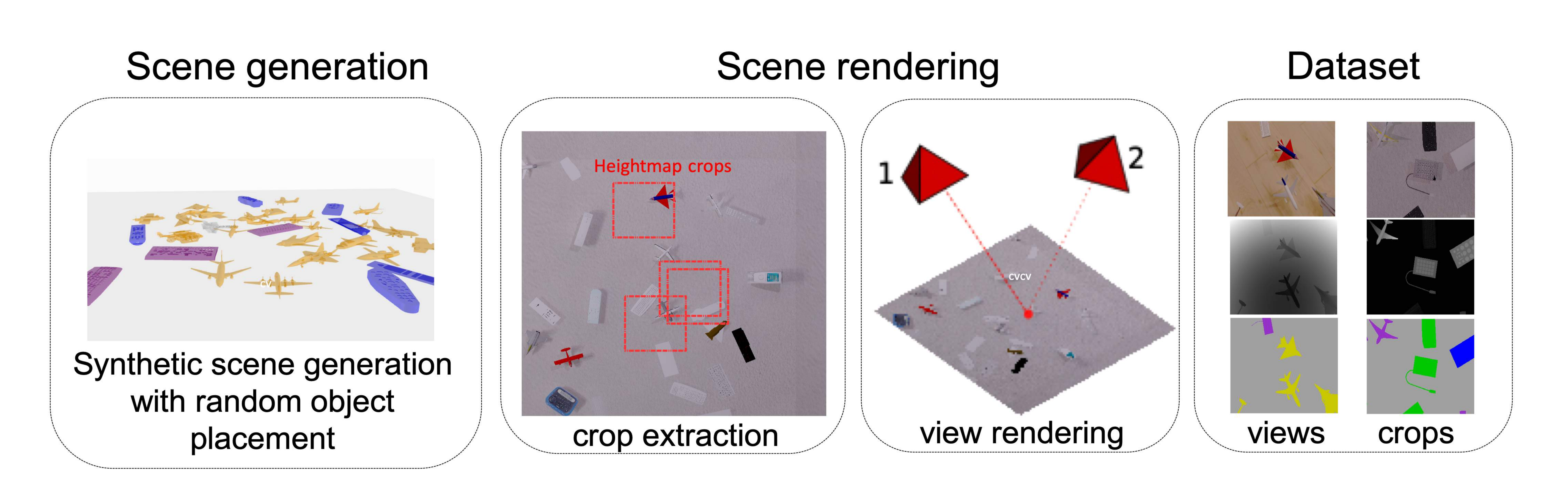}
\end{center}
  \caption{Dataset generation. A random synthetic scene (\textbf{left}) from which a height map, height map-crops and views are extracted (\textbf{centre}), yielding our datasets of views (RGB, depth, semantic segmentation) and height map crops (RGB, height, semantic segmentation) (\textbf{right}).}
\label{fig:dataset}
\vspace{2mm} \hrule
\end{figure}

We created 647 scenes composed of objects selected from the \textit{ShapeNet} taxonomy \cite{chang2015shapenet} of the object categories \textit{computer keyboard, keypad}, \textit{remote control, remote} and \textit{airplane, aeroplane, plane}, which are suitable for a height map representation as they typically have little overhang. For each category, we select $66$ instances, split into training, validation and test data with fractions $75\%, 10\%$ and $15\%$ respectively. 
We chose SceneNet RGB-D \cite{McCormac:etal:ICCV2017} for its photorealistic rendering and adapt its rendering engine to render height maps by simulating an orthogonal camera using the OptiX Raytracing engine. We generate random scenes by selecting and placing instances from each object category, sampling from a uniform distribution for object ID, $(x,y)$ position and orientation. We generate random backgrounds to increase variability in appearance. We then create data to train our view-based and map-based labelling approaches. For the former, we extract scene views ($240\times320$) at random camera locations with RGB, depth and semantic ground truth. For the latter, we use height map fusion with ground truth camera poses to reconstruct height maps ($1025\times1025$) for which we obtain semantic ground truth by rendering the same scene in SceneNet RGB-D \cite{McCormac:etal:ICCV2017}. We then extract map samples of ($240\times320$) at random locations. Note that while we vary the camera height between $0.18m$ and $0.4m$ and extract views with large angle variance ($0\degree - 40 \degree$), the rendered height maps exhibit canonical scale and orientation. We generate $500$ training and $120$ validation scenes from which we use views and height map crops to train our view-based and map-based networks respectively. From our $27$ test scenes, we use views and crops to evaluate network performance and use entire scenes in our evaluation pipeline (see Section \ref{sec:evaluation_pipeline}) to compare both methods during scene reconstruction. An example of our synthetic dataset is shown in Figure \ref{fig:dataset}.

\subsection{Network Architecture and Training}
\begin{figure}[t]
\centering
\begin{center}
  \includegraphics[width=1\linewidth]{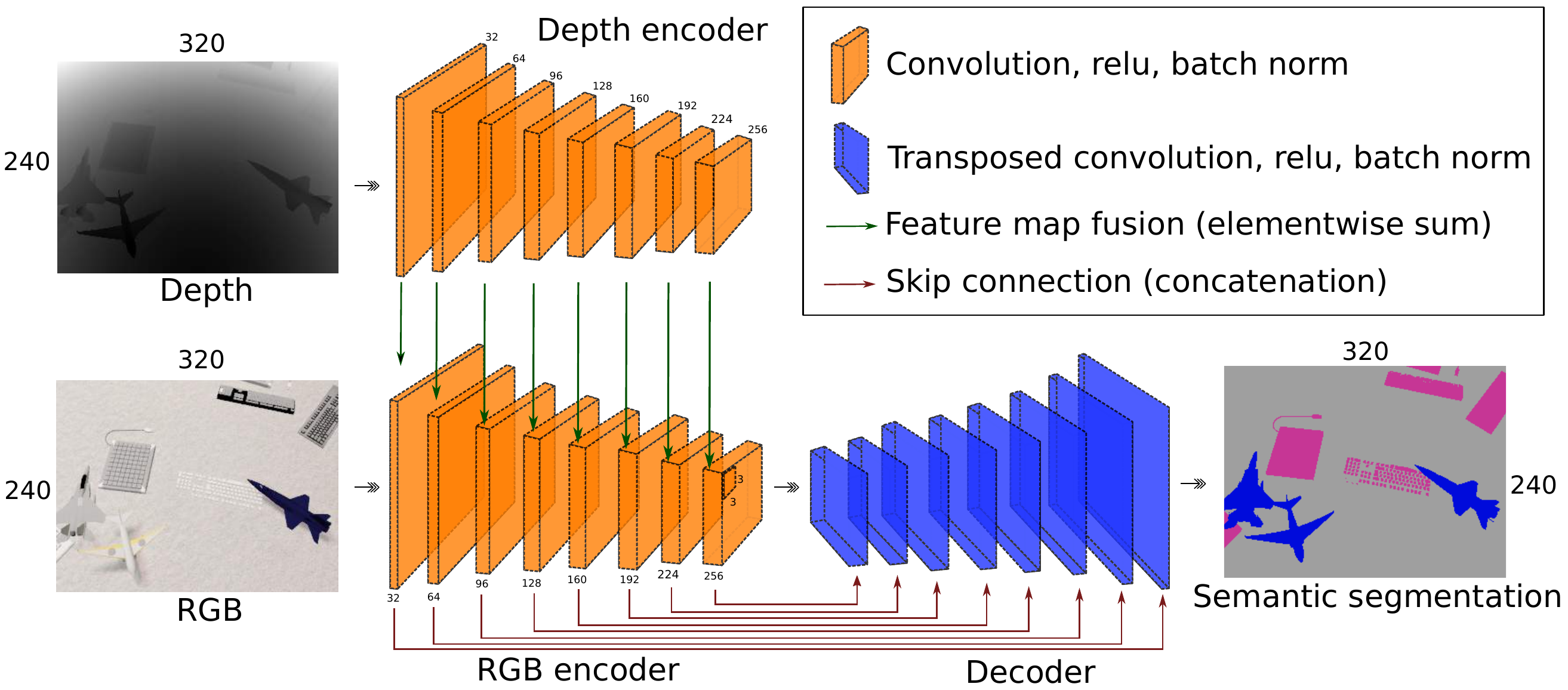}
\end{center}
  \caption{Illustration of the fully convolutional architecture of the networks we use for view-based and map-based labelling. Depth (view-based) or height (map-based) and RGB encodings are fused at every encoding layer. Both encoders consist of 8 convolutional modules. Deconvolutional modules consist of one upsampling and two convolutional layers.}
\label{fig:network_architecture}
\vspace{2mm} \hrule
\end{figure}

We use a fully convolutional network based on the \textit{Fusenet} architecture~\cite{hazirbas2016fusenet}, which uses two parallel encoder branches (one for RGB, one for depth), to predict semantic labels for every pixel (see Figure \ref{fig:network_architecture}). We  experiment with different network architectures to obtain the best performing network with a minimum number of parameters. The best performing model was trained with a batch size of $8$, a drop-out rate of $0.1$ and a learning rate of $0.001$ with exponential decay of base $0.96$ every $1e^5$ steps. All our models were developed with Tensorflow using the Adam optimizer \cite{kingma2014adam}. 


\subsection{Evaluation and Comparison}
\label{sec:evaluation_pipeline}
We evaluate and compare both approaches using our extension of \textit{Semantic Height Map Fusion}~\cite{zienkiewicz2016monocular}. We generate test sequences from our test scenes which have camera locations relative to the scene, randomly sampled from a range which stochastically achieves full scene coverage. 
For each test sequence, we deploy the view-based network during the reconstruction of a synthetic scene and use our Bayesian Fusion update scheme to obtain a semantically labelled height map. We save the reconstruction at regular intervals where it is labelled by the map-based network. The semantic scene segmentation obtained by each approach is compared against the scene's ground truth using the mean Intersection over Union ($IoU$) over all classes. 

\section{Experiments and Results}
\label{sec:experiments}
\subsection{Training results}
\begin{figure}[t]
\begin{center}
  \includegraphics[width=1.0\linewidth]{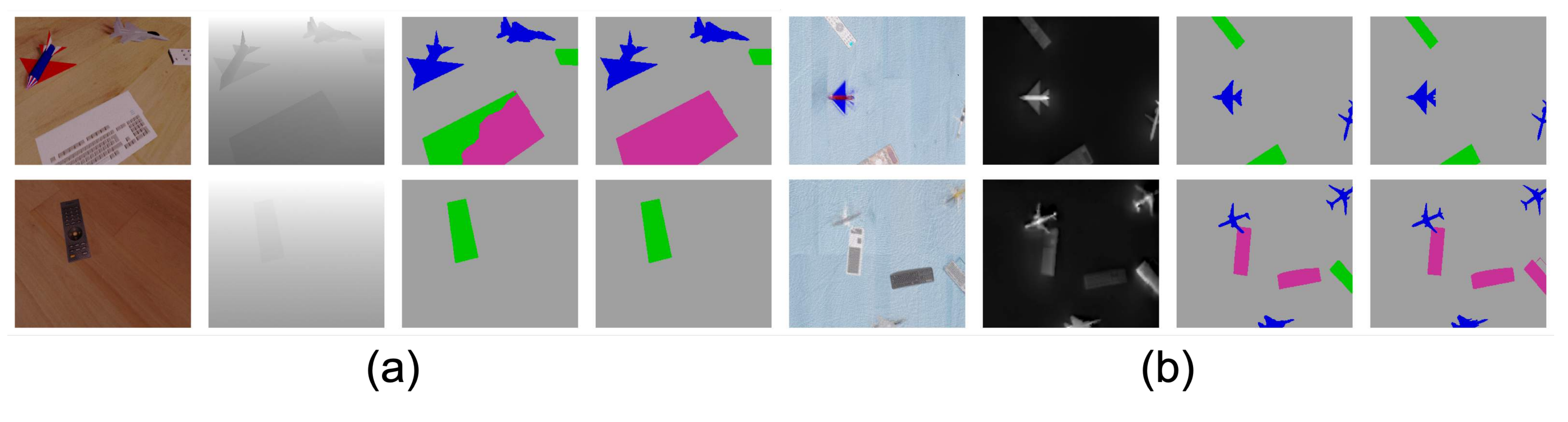}
\end{center}
  \caption{Qualitative results from the view-based network (\textbf{a}) and the map-based network (\textbf{b}). Left to right: Input RGB, input depth (view-based) or input height (map-based), network output, and ground truth. The image dimensions are $320\times240$.}
\label{fig:qualitative_val_perf}
\vspace{1.1mm} \hrule
\end{figure}

Using our best architecture (8 convolutional layers and skip layers at every downsampling step), we achieved a mean IoU of $0.95$ and $0.93$ for the view-based and map-based tasks respectively (see Table \ref{tab:validation_performance}). We suggest that this discrepancy in accuracy, occurring despite identical architecture and training sample number, is due to the different characteristics of the data seen by the networks, arising from the nature of their tasks and the distributions of their views. Compared to the view-based task, the map-based task is easier to learn since all map crops are taken from a canonical top-down orientation of the camera. The lower variability can possibly also lead to stronger overfitting on the training data and could explain the lower performance of our map-based method on the test data. Qualitative results for both networks can be seen in Figure \ref{fig:qualitative_val_perf}.

\begin{table}[t]
\footnotesize
\begin{center}
\begin{tabular}{|l || c || c | c | c | c |}
\hline
 & mIoU & surface & remote control & keyboard & model plane \\
\hline\hline
Vb & 0.95 \% &0.99 \% &0.89 \% &0.74 \% & 0.51 \% \\ \hline  
Mb & 0.93 \% &0.99 \% &0.68 \% & 0.91 \% & 0.78 \%  \\
 \hline
\end{tabular}
\end{center}
\caption{Performance of our view-based (Vb) and map-based (Mb) network on the test dataset of single views and map crops, respectively. Evaluated using mean Intersection over Union.}
\label{tab:validation_performance}
\vspace{1.1mm} \hrule
\end{table}

\subsection{Evaluation of the View-Based method}
We evaluate our view-based method on our test scenes by generating a semantic height map using our semantic fusion algorithm as described in \ref{sec:evaluation_pipeline}. We evaluate at every 100 frames, to track the increasing semantic accuracy of the reconstructed scene. On average, our view-based method reaches a value of $0.889$ mean IoU after $1000$ frames. Figure \ref{fig:view_and_map_based_results} (left) displays our results for all test scenes, together with the average performance. As illustrated by the semantic reconstruction example at different coverage levels (number of seen frames) in Figure \ref{fig:teaser}, the rapid early improvement in label accuracy is obtained from increasing coverage of the map, though the IoU continues to improve slowly due to fusion once full coverage has been achieved at around 400 frames. This is not surprising given the high labelling performance of the network on single images. With a more poorly performing network we would expect an even higher increase in accuracy from incremental fusion. We tested the influence of the alpha parameter in Equation \ref{eq:distance_based_decay} on our semantic reconstruction. It did not affect the reconstruction strongly in this setting and we chose a value of $1.0$ for our experiments. We further performed an analysis of computational time for our view-based method, presented in Table \ref{tab:computational_analysis}.

\begin{table}[t]
\footnotesize
\begin{center}
\begin{tabular}{|l | c|}
\hline
     Data loading & 6.13 ms \\
     \hline
     Reconstruction & 1.78 ms \\
     \hline
     Semantic segmentation (1 forward pass) & 77.00 ms \\
     \hline
     Semantic fusion & 31.05 ms \\
     \hline

\end{tabular}
\end{center}
\caption{Computational analysis (per frame average) of our view based method for $27$ synthetic test scenes reconstructed from $1000$ frames.}
\label{tab:computational_analysis}
\vspace{1.3mm} \hrule
\end{table}

\subsection{Evaluation of the Map-Based Method}

\begin{figure}[t!]
\begin{center}

  \includegraphics[width=\linewidth]{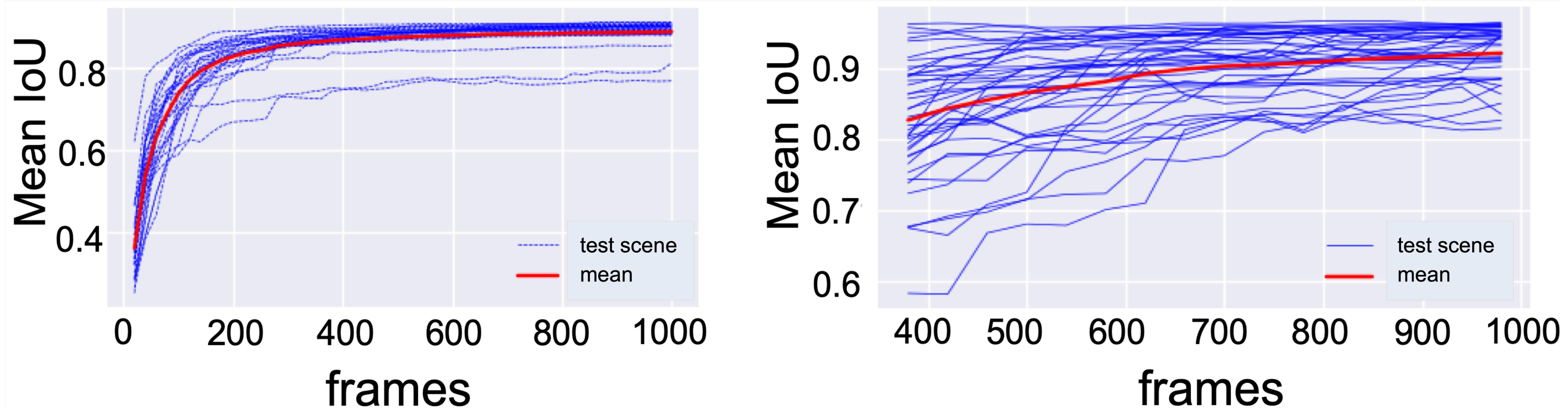}
\end{center}
  \caption{Incremental fusion results (\textbf{left}) and map-based results (\textbf{right}). For both approaches the semantic labelling accuracy is measured as the mean IoU over all classes for $27$ scenes. The red line shows the average over all test scenes which reaches a maximum of $\textbf{0.889}$ and $\textbf{0.922}$ at 1000 seen frames for the view-based and the map-based, respectively.}
\label{fig:view_and_map_based_results}
\vspace{2mm} \hrule
\end{figure}

We evaluate our map-based method using the same test sequences. We reconstruct the scene geometry using variable number of frames (up to $1000$) and segment the reconstructed height maps using the sliding window method with a shift set by the theoretical receptive field of $(91,91)$ of our network. Unlike the view-based method, we start the map-based evaluation once full coverage of the scene has been reached, to avoid segmenting incomplete image patches which lie outside of the network's learned distribution. Figure \ref{fig:view_and_map_based_results} (right) shows our results over all scenes. Our map-based method achieves an average mean IoU of $0.922$. We measure the average time to segment one reconstructed map on a single GPU as $8.3s$.

\subsection{Both approaches in comparison}

\begin{figure}[t!]
\begin{center}
  \includegraphics[width=\linewidth]{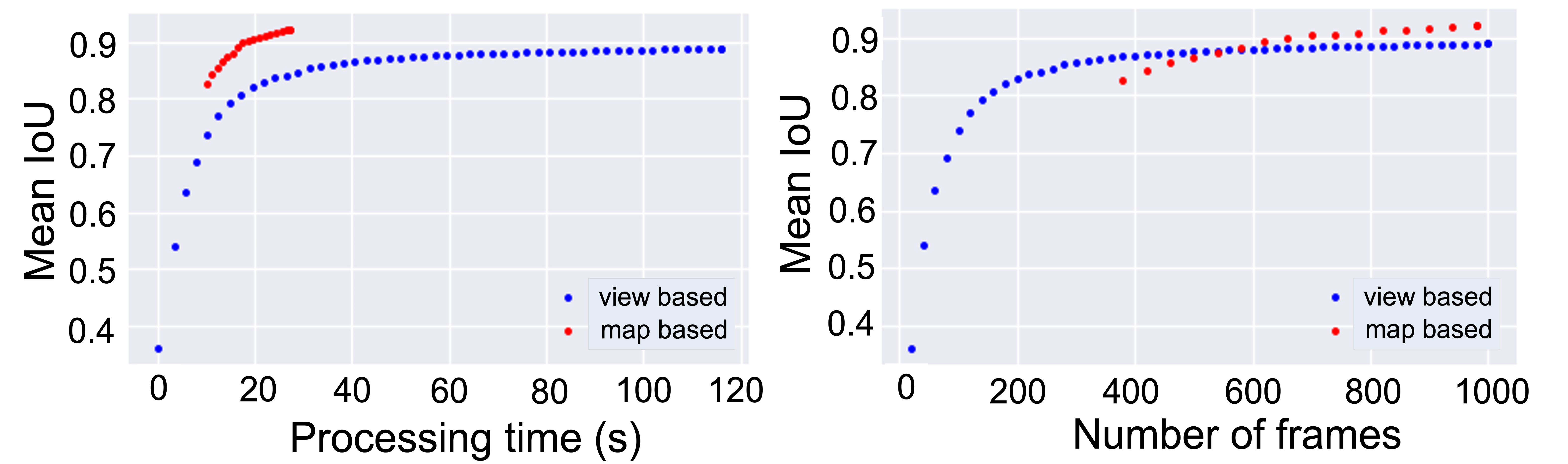}
\end{center}
  \caption{Semantic labelling accuracy achieved by the view-based method and the map-based method in comparison. \textbf{Left}: The average of the mean IoU over all classes for $27$ test scenes, plotted against the number of frames used in the reconstruction. \textbf{Right}: The same results plotted against processing time. For the view-based method, we measure the time for reconstruction, the network forward pass at every keyframe and semantic fusion. For the map-based method, we combine the processing time for reconstruction without semantic fusion with the processing time required to segment the map.}
\label{fig:results_mean_iou_comparison}
\vspace{2mm} \hrule
\end{figure}

We compare the mean IoU achieved on average by both methods on our test scenes during reconstruction, evaluating at every $20$ and $40$ frames for the view-based and for the map-based method respectively, demonstrating each method's improvement w.r.t. the reconstruction state. We also compare both methods with regards to processing time to evaluate their efficiency. The results are shown in Figure \ref{fig:results_mean_iou_comparison}. For the view-based method, the computation time of the reconstruction, frame-based semantic segmentation and semantic fusion is measured, and for the map-based approach, we measure the computation time for the reconstruction and the one-off scene labelling. Note that the overall processing time of our map-based method could be reduced further, if we had only segmented the map once after reconstructing the full scene. Our results show that on average, with overall much less computation time, the map-based method achieves a segmentation accuracy superior to the view-based method. However, after full coverage has been reached, we observe a region in which the view-based method achieves higher labelling accuracy than the map-based method (see Figure \ref{fig:results_mean_iou_comparison} on the right). This demonstrates that for the map-based method to work well, a certain level of reconstruction accuracy has to be reached. We further observe that the map based method performs better at the contours of objects than the view-based method. This is visualised in the error maps of the reconstruction example (Figure \ref{fig:teaser}).

\subsection{Comparison with reduced map reconstruction quality}
\begin{figure}[t!]
  \includegraphics[width=\linewidth]{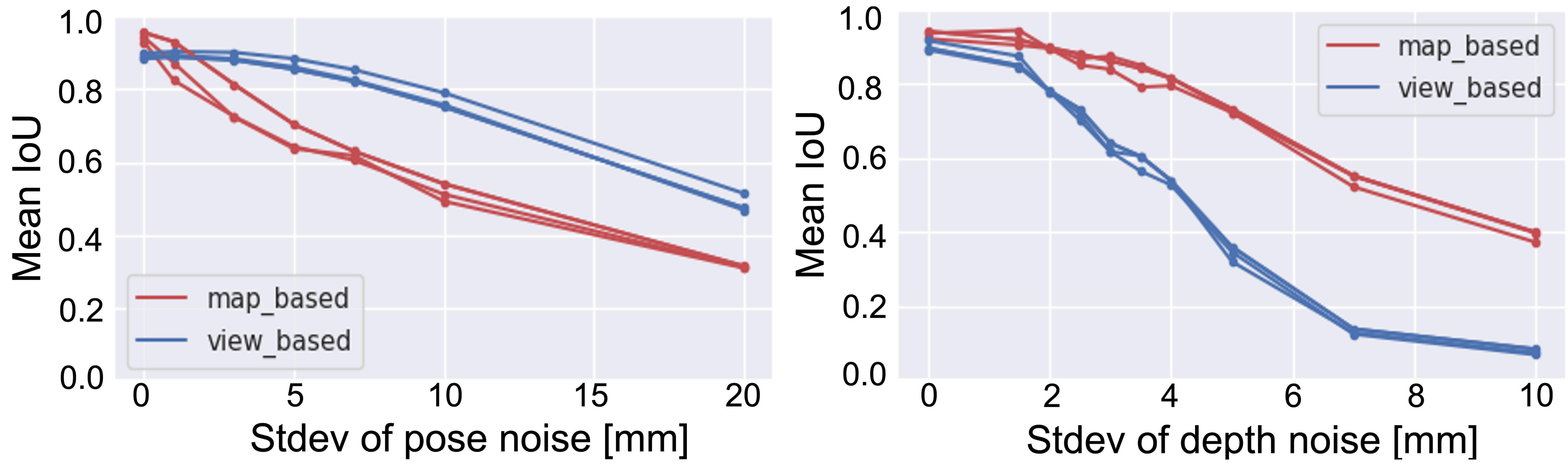}
   \caption{Semantic labelling accuracy over $3$ randomly selected test scenes with increasing noise (\textbf{left}: pose noise, \textbf{right}: depth noise) We apply normally distributed pose and (per pixel) depth disturbances.}
\label{fig:results_noise_study}
\vspace{2mm} \hrule
\end{figure}

\begin{figure}[t!]
\begin{minipage}[c]{0.49\linewidth}
  \includegraphics[width=\linewidth]{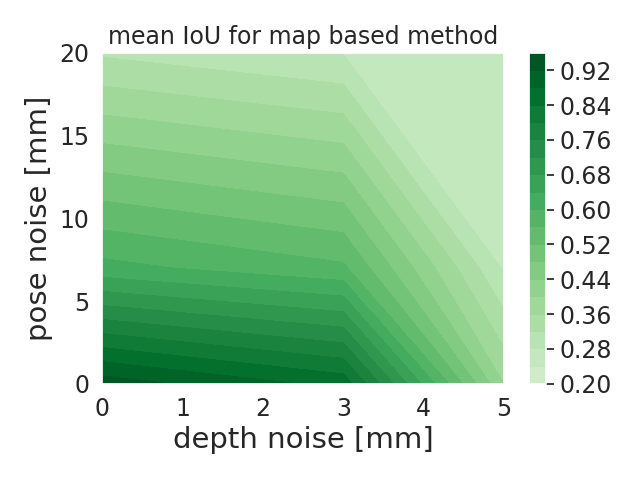}
\end{minipage}
\begin{minipage}[c]{0.49\linewidth}
  \includegraphics[width=\linewidth]{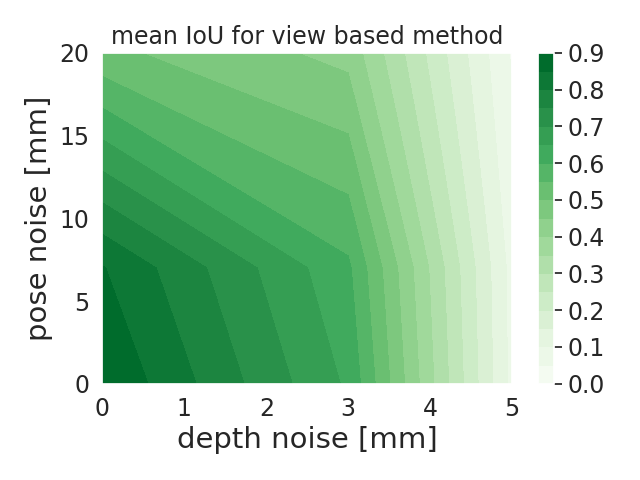}
\end{minipage}
   \caption{Map-based (\textbf{left}) and view-based (\textbf{right}) mean IoU as a function of pose and depth noise. Darker regions indicate higher labelling performance for each method.}
\label{fig:results_noise_study2}
\vspace{2mm} \hrule
\end{figure}

We experiment with degrading the map quality to evaluate the decrease in labelling quality w.r.t. noise, on the compared methods. In a first study, we apply normally distributed pose disturbances during reconstruction. Our results (Figure \ref{fig:results_noise_study}) show that the map-based method is much less robust to pose noise than the view-based method, which can be attributed to the Bayesian filtering of multiple views. We then apply depth noise drawn from a normal distribution which has a stronger negative effect on the view-based method, most likely because the latter now has to deal with reduced quality in 2D labels as well as projection errors. Plotting both methods against pose and depth noise (Figure \ref{fig:results_noise_study2}) shows that while the view-based method is robust to pose noise, it quickly degrades to $<0.2$ mean IoU with increasing depth noise. On the other hand, the map-based method is sensitive to both noise types, but to a lesser degree, degrading to only $0.2$ mean IoU in the tested noise range.


\section{Discussion}
Our results show that for a setting which assumes perfect poses and depth, the map-based approach achieves higher labelling accuracy, and can be achieved with less computation. We argue that although the view-based network achieves a high labelling accuracy on individual frames, its deployment during the early reconstruction phase results in more errors, especially in border regions of objects, from which it cannot always recover easily. The overhead of repeated forward passes and multi-view fusion is a further disadvantage of this approach. Our experiments on pose and depth noise show that for the studied noise range, the map-based approach, although more equally affected by both types of noise, stays overall more robust than the view-based method (Figure \ref{fig:results_noise_study2}). We argue that its advantage results from operating on fused data. The view-based method on the other hand is less affected by pose noise, due to its in-view labelling and the benefits of Bayesian label fusion. However, it strongly degrades in the presence of depth noise. We argue that this is caused by the fact that it has to operate directly on the noisy depth data, resulting in not only projection errors, but also 2D segmentation errors.

Overall, the present results should be carefully considered within the context of our experiments. Firstly, the 2.5D geometry of the height map alleviates the map-based method from the memory limitations of 3D segmentation tasks. Secondly, the appearance features of our selection of scattered objects with little overhang are well visible from a top-down perspective, while for other objects (e.g. cups, bowls, chairs) with more ambiguous features it would be more difficult to train a top-down network. In settings with less well performing networks, one-off labelling of the map segments would likely yield more errors which would in turn be better recovered with the view-based method due to multi-view Bayesian fusion of labels. While we don't cover the cases of differently well performing networks in this study, we would like to leave it for future work.


\section{Conclusion}
We conclude that in the absence of noisy data, the map-based approach shows higher labelling accuracy and object-border details. In the presence of pose noise, only affecting the reconstruction, the view-based method shows a significant advantage, but it deteriorates more strongly in the presence of depth noise. In terms of computational cost, the map-based approach is in principle more efficient, given that every map element is processed only once, compared to the frame-wise segmentation and fusion required by the view-based method. In a three dimensional setting, this advantage will likely be reduced due to expensive 3D data processing such as volumetric labelling, but we leave this analysis for future work, as it requires the comparison between different representations of 3D data and their different methods of 3D labelling. We see a further point of continuation in combining both view-based and map-based methods, leveraging the respective advantages in the correct setting. For instance, a real-time SLAM system would benefit from incremental fusion in the initial mapping phase and a regular map-based label refinement step in well-reconstructed regions of the map.

\small
\bibliographystyle{IEEEtran}
\bibliography{IEEEabrv,robot_vision_bib_}

\end{document}